\documentclass[runningheads]{llncs}
\usepackage[T1]{fontenc}
\usepackage{graphicx}
\usepackage{amsmath,amssymb}
\usepackage{booktabs,multirow,tabularx,array}
\usepackage{enumitem}
\usepackage{adjustbox}
\usepackage{microtype}
\usepackage{url}
\graphicspath{{figures/}}

\begin{document}
\title{QFacePlacement: From Arbitrary Tensorization to Semantic Spatial Correspondence in Modern Face Verification}
\titlerunning{QFacePlacement}

\author{Krishna Bhatia\inst{1}, Kavita Choudhary\inst{2}, Pruthwik Mishra\inst{2}, and Praveen Kumar Chandaliya*\inst{2}}
\authorrunning{Bhatia et. al}
\institute{QuantumAI Lab, Fractal Analytics, Mumbai, India\\\email{krishna.bhatia@fractal.ai}
\and Department of Artificial Intelligence, Sardar Vallabhbhai National Institute of Technology, Surat, India \\\email{ds25ai002@aid.svnit.ac.in,pruthwikmishra@aid.svnit.ac.in,pkc@aid.svnit.ac.in*}}

\maketitle

\begin{abstract}
Quantum-inspired computer vision often begins by reshaping a high-dimensional representation into a tensor and then attaching a tensor-network model to it. In face verification this can be misleading: modern embeddings are already optimized for a simple cosine geometry, and a formal tensorization does not guarantee the low-entanglement structure required by matrix-product operators or tensor trains. We study \emph{placement} through two controlled interfaces that use the same modern face-recognition setting but expose very different mathematical objects. First, we examine cross-model embedding compatibility, in which a 512$\times$512 alignment operator must map AdaFace embeddings into the ArcFace space. Dense Ridge and Procrustes maps remain strong, whereas deterministic TT-SVD/MPO compression fails even after basis transformations and low-rank residual hybrids: at the final 64K storage budget, dense Ridge averages 0.9297 directional AUC, rank-64 SVD 0.8952, the best low-rank-MPO hybrid 0.8685, and raw MPO approximately 0.649. Second, we move before global pooling and construct a genuine $7\times7\times7\times7$ probe--gallery correspondence tensor from local AdaFace descriptors. Here a geometry-aware symmetric TT with bond dimension 2 uses only 122 trainable parameters versus 2403 for a dense linear matcher and preserves LFW AUC (0.9802), showing that semantically meaningful tensor structure enables strong compression. However, it does not improve the stringent biometric tail: on CFP-FP, TAR at FAR=$10^{-3}$ falls to 0.0102 versus 0.1734 for global cosine, and on CPLFW it reaches 0.7805 versus 0.8737. The resulting principle is sharper than ``tensor networks help'' or ``tensor networks fail'': quantum-inspired factorization is credible only when the chosen interface exposes task-aligned axes and compressible structure, and aggregate AUC must be checked against operational low-FAR behaviour. This placement-oriented analysis provides a reusable design doctrine for quantum computer vision and biometric matching.
\end{abstract}

\section{Introduction}
Modern face verification is a difficult test bed for adding a new learning primitive. FaceNet~\cite{schroff2015facenet}, CosFace~\cite{wang2018cosface}, ArcFace~\cite{deng2019arcface}, and AdaFace~\cite{kim2022adaface} deliberately organize identity representations so that normalized cosine similarity already provides an effective verifier. Consequently, additional model capacity is useful only if the new module sees information or structure that the pooled embedding does not already expose. This observation is especially important for quantum and quantum-inspired computer vision. Quantum kernels and variational circuits require aggressive state preparation or dimensional compression, while tensor-network (TN) models require a useful factorization of the object on which they operate. A vector of length 512 can always be reshaped because $512=2^9$, but algebraic reshaping alone does not imply that the resulting nine-site object has low operator entanglement or short-range dependencies. Conversely, a local correspondence tensor whose indices explicitly denote probe and gallery coordinates has a semantic product structure before any model is introduced.

Quantum face-recognition work spans multigate QCNNs~\cite{zhu2024mgqcnn}, quantum-enhanced eigenface/PCA formulations~\cite{singh2025eigenface,singh2025qpca}, quantum optical acquisition~\cite{salari2023ghost}, and quantum-inspired attention~\cite{gauba2026attention}. Our earlier near-term study likewise found that compact quantum and tensorized modules can be useful under restricted data or resource budgets while strong classical face representations remain difficult to displace~\cite{bhatia2026nearterm}. These results motivate a question that precedes architecture choice: \emph{which interface actually has the structure a quantum-inspired factorization needs?}.

\begin{figure}[t]
\centering
\includegraphics[width=\textwidth]{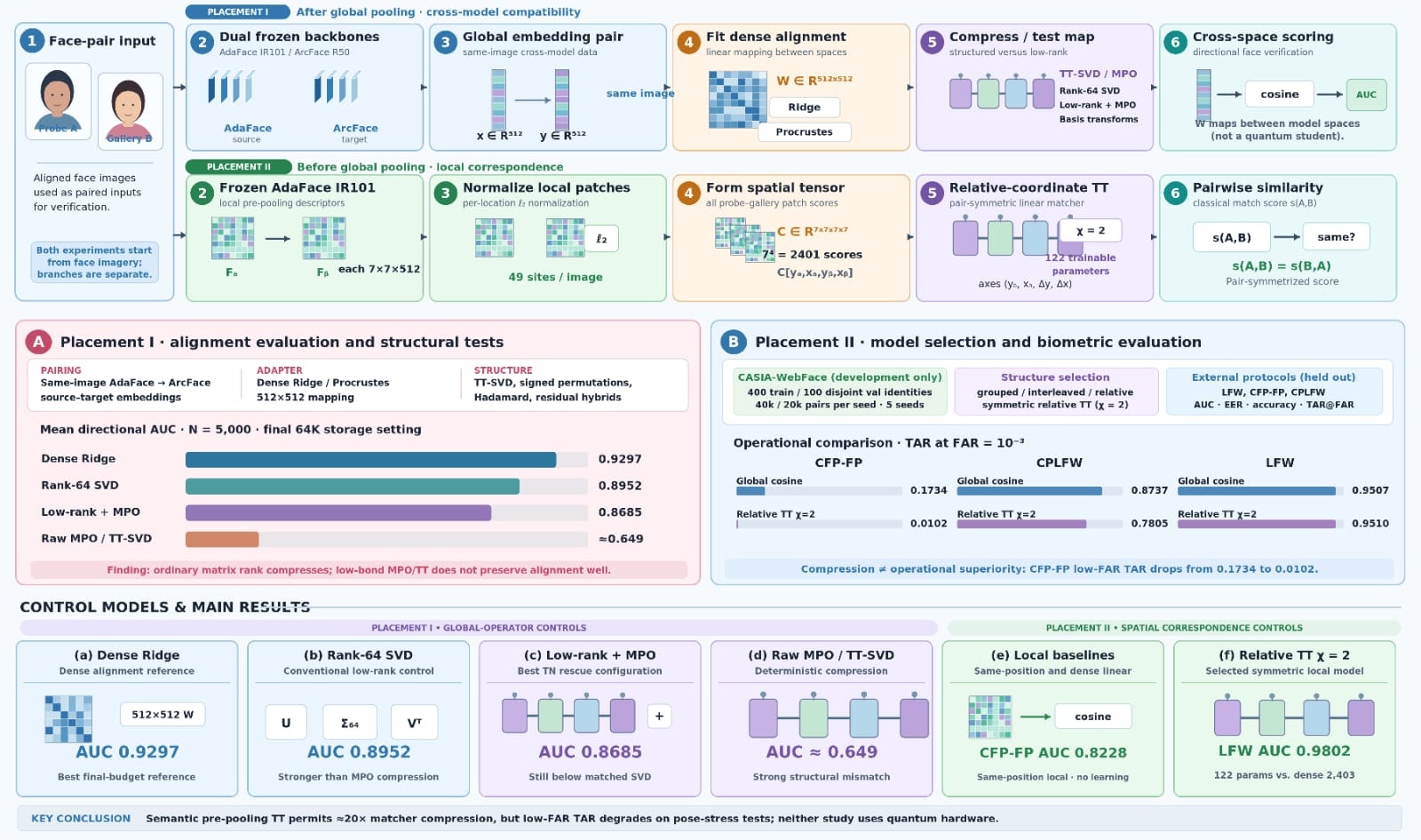}
\caption{QFacePlacement contrasts two interfaces. Compatibility alignment tensorizes an already-global embedding operator; local correspondence acts before global pooling on axes with explicit probe/gallery geometry. The same TN intuition is therefore tested on mathematically different objects.}
\label{fig:overview}
\end{figure}

We answer this question through two experiments that deliberately contrast \emph{arbitrary} and \emph{semantic} tensorization (Fig.~\ref{fig:overview}). The first studies cross-model compatibility, where the object is a dense 512$\times$512 linear operator between independently trained face-embedding spaces. The second moves before global pooling and operates on the full spatial similarity tensor between two $7\times7$ local descriptor grids. Both studies retain strong dense and low-rank classical controls and use external biometric protocols rather than random image splits.

Our main contributions are:
\begin{itemize}[leftmargin=*]
\item We formulate semantic structure at the insertion point as the experimental variable, rather than selecting a Tensor Network (TN) architecture first and searching for an input shape.
\item We show that a dense face-compatibility operator remains poorly represented by low-bond MPO/TT structure even under deterministic TT-SVD, basis transformations, and low-rank residual hybrids. A storage-matched truncated SVD is consistently stronger, indicating ordinary matrix-rank compressibility rather than TN locality.
\item We show that a true local correspondence tensor is a substantially better structural fit: a relative-coordinate symmetric TT with only 122 parameters reaches 0.9802 LFW AUC and competitive stress-set AUC while providing roughly 20$\times$ compression relative to dense matching.
\item We show that parameter efficiency and AUC are not sufficient biometric claims. On the two pose-stress protocols, the learned TT reshapes the impostor tail unfavorably and loses substantial TAR at FAR=$10^{-3}$.
\item We derive a placement doctrine for quantum computer vision: identify semantically factorized interfaces, measure whether the target operator is actually compressible under the proposed structure, compare against simple low-rank controls, and evaluate operational tails before attributing value to a quantum-inspired model.
\end{itemize}

\section{Related Work}
\subsection{Modern Face Verification and Local Evidence}
Face recognition has progressively moved from metric-learning objectives to highly discriminative hyperspherical embeddings. FaceNet established triplet supervision~\cite{schroff2015facenet}; ArcFace and CosFace improved angular separation~\cite{deng2019arcface,wang2018cosface}; AdaFace adapts margins to recognizability~\cite{kim2022adaface}; and MagFace links embedding magnitude with face quality~\cite{meng2021magface}. These approaches make it difficult to improve a global cosine score indiscriminately. Nevertheless, global pooling is not guaranteed to retain every cue useful under pose, occlusion, or domain shift. DeepFace-EMD, for example, operates on local descriptors and uses patch-wise optimal transport for out-of-distribution matching~\cite{phan2022deepfaceemd}. This distinction motivates our second placement: if local descriptors still retain spatially organized identity evidence, a structured matcher should be tested before that information is pooled away.


\subsection{Quantum and Quantum-Inspired Vision}

Quantum feature maps construct learned or fixed representations in Hilbert space~\cite{schuld2019featurehilbert,havlicek2019supervised}, while data re-uploading can increase the expressive power of shallow parameterized circuits~\cite{perezsalinas2020datareuploading}. However, near-term quantum models remain constrained by state preparation, limited qubit counts, measurement overhead, and trainability issues such as barren plateaus~\cite{mcclean2018barren}. These constraints are particularly relevant to face recognition, where strong pretrained embeddings commonly contain hundreds of dimensions.

Tensor networks (TNs) provide a complementary quantum-inspired approach by representing high-order tensors through structured low-rank factorizations. Tensor Trains (TTs) and Matrix Product States (MPS) offer compact representations of high-dimensional arrays~\cite{oseledets2011tt}, while tensorized neural networks can substantially reduce the parameterization of dense layers~\cite{novikov2015tensorizing}. MPS-based models have also demonstrated supervised learning directly on product-structured feature representations~\cite{stoudenmire2016supervised}. Crucially, however, the effectiveness of TN compression depends on whether the target representation exhibits correlations that can be captured at a manageable bond dimension. We therefore treat this structural assumption as an empirical hypothesis and explicitly test whether the location and semantics of tensorization determine the effectiveness of TN representations for biometric matching.
\section{Why Placement Is a Structural Question} The central motivation of this work is that the effectiveness of a Tensor Network (TN) representation depends not only on the chosen architecture, but also on \emph{where and how} tensorization is introduced. A compact representation can arise for fundamentally different reasons. First, the underlying operator may possess low conventional matrix rank, in which case truncated SVD provides a natural and often highly competitive compression mechanism. Second, after a meaningful tensorization, the operator may exhibit low Schmidt rank across successive tensor bipartitions, making an MPO/TT representation structurally appropriate. These two notions of compressibility are distinct: low matrix rank does not imply TN locality, and a compact TN representation should not be credited with an advantage that is already explained by ordinary low-rank structure.

This distinction motivates our experimental design. Rather than selecting a TN architecture first and adapting the data to its required tensor shape, we treat \emph{tensorization placement and semantic structure} as the primary experimental variables. We examine both a dense cross-model compatibility operator and a genuinely local correspondence tensor, thereby testing whether TN compression is favored by the structure of the object being represented. For each structured representation, we compare against dense, truncated-SVD, CP/Tucker, and parameter-free controls under appropriate storage and evaluation constraints. This design isolates whether observed gains arise from genuine TN-compatible structure or merely from generic low-rank compression, and establishes placement as a structural question rather than an architectural convenience.

\section{Placement I: Cross-Model Embedding Compatibility}
\subsection{Problem Definition} Model upgrades create a practical compatibility problem: a gallery encoded by a legacy face model may remain deployed while new probes are produced by a newer model. We use AdaFace IR101 trained on WebFace12M as the source representation and ArcFace R50 as the target. Given same-image source--target pairs $(x_i,y_i)\in\mathbb{R}^{512}\times\mathbb{R}^{512}$, the goal is to learn $f(x)\approx y$ such that probe-to-gallery verification remains meaningful across model spaces. Dense baselines include linear least squares, Ridge regression, and orthogonal Procrustes~\cite{schonemann1966procrustes}. The structured family includes butterfly orthogonal maps, MPO/TT adapters, local unitary TNs, and spectral-unitary TNs. The decisive salvage experiment starts from the fitted dense Ridge operator and applies deterministic TT-SVD, eliminating gradient optimization as a confound. We additionally test signed permutations, Walsh--Hadamard basis changes, and low-rank-plus-MPO hybrids.
\begin{table}[t]
\centering
\caption{Cross-model compatibility at $N=5000$ and the final 64k storage budget. Values are mean directional verification AUC}
\label{tab:compat}
\begin{tabular}{lcc}
\toprule
\textbf{Adapter} & \textbf{Mean AUC} $\uparrow$ & \textbf{Interpretation}\\
\midrule
Dense Ridge & \textbf{0.9297} & Reference map\\
Rank-64 SVD & 0.8952 & Strongest compact control\\
Low-rank + MPO hybrid & 0.8685 & Best TN salvage\\
Raw MPO / TT-SVD & $\sim$0.649 & Severe mismatch\\
\bottomrule
\end{tabular}
\end{table}



\begin{figure}[t]
\centering
\includegraphics[width=.82\textwidth]{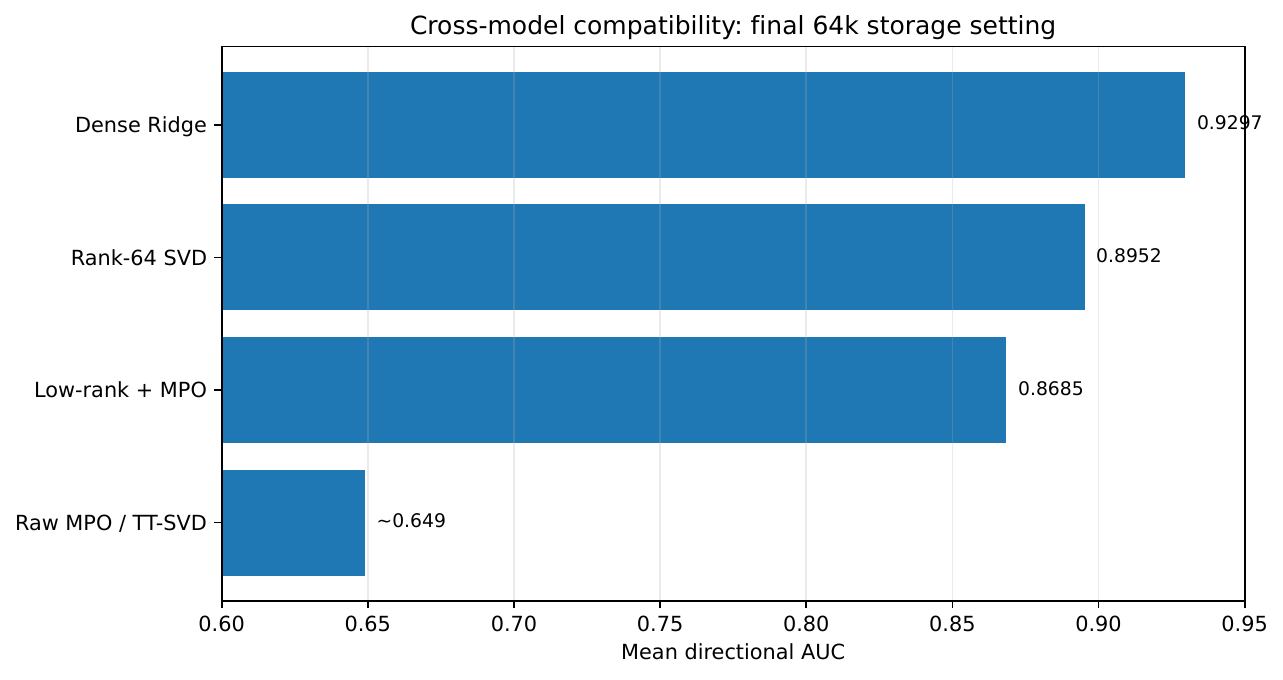}
\caption{Final compatibility salvage. Conventional truncated SVD retains substantially more verification structure than the tested MPO/TT factorization at the same broad storage scale.}
\label{fig:compat}
\end{figure}
\subsection{Storage-Matched Evaluation}
We adopt a deliberately stringent, pre-registered success criterion: a structured candidate must (i) remain within 0.01 mean directional AUC of dense Ridge, (ii) remain within 0.002 AUC of a storage-matched truncated-SVD baseline, and (iii) require no more than 25\% of the storage used by the dense operator. Under this criterion, a TN representation is considered successful only if it captures structure that cannot be explained by conventional low-rank compression at a comparable storage budget.
The results in Table~\ref{tab:compat} and Fig.~\ref{fig:compat} show that the dense cross-model compatibility operator does not satisfy this requirement. Dense Ridge achieves a mean directional AUC of 0.9297, while rank-64 SVD retains 0.8952. In contrast, the strongest low-rank--MPO hybrid reaches only 0.8685, and the raw MPO performs substantially worse, with an AUC of approximately 0.649. The same ordering is observed under the earlier symmetric evaluation: Procrustes achieves AUCs of 0.980/0.828/0.965/0.979 on LFW/CFP/CASIA/VGGFace2, closely matched by Ridge at 0.980/0.827/0.965/0.978. By comparison, an MPO with bond dimension 8 achieves only 0.737/0.569/0.729/0.719, while the compact spectral TN obtains 0.668/0.585/0.646/0.632.

The structural diagnostics are consistent with this performance gap. Operator--Schmidt entropy remains high across the important tensor bipartitions, providing no evidence that the dense alignment operator admits a strongly localized low-bond representation in the tested factorizations. Importantly, this does not imply that the operator is incompressible: the SVD result demonstrates meaningful conventional matrix-rank compressibility. Rather, the evidence indicates that its useful compressibility is predominantly \emph{global matrix-rank structure}, not the successive low-Schmidt-rank structure required for an efficient MPO/TT factorization. Thus, for this dense compatibility operator, tensor-network compression introduces an unfavorable structural constraint rather than exposing additional compressibility beyond what a conventional low-rank approximation already provides.

\section{Placement II: Pre-Pooling Local Correspondence}


\subsection{Semantically Factorized Tensor}

We next move the tensorization point before global pooling, where the spatial structure of the two images is still explicit. A frozen AdaFace IR101 backbone produces a $7\times7$ spatial grid of local $512$-dimensional descriptors. Let
$
F_A,F_B\in\mathbb{R}^{7\times7\times512}
$
denote the feature grids for probe image $A$ and gallery image $B$, respectively. After $\ell_2$ normalization of each local descriptor, we construct the pairwise local-similarity tensor
\begin{equation}
C_{y_A,x_A,y_B,x_B}
=
\left\langle
\widehat{F}_A[y_A,x_A],
\widehat{F}_B[y_B,x_B]
\right\rangle,
\qquad
C\in\mathbb{R}^{7\times7\times7\times7},
\label{eq:local_similarity_tensor}
\end{equation}
where $y_A,x_A,y_B,x_B\in\{1,\ldots,7\}$. Thus, $C$ contains $7^4=2401$ local correspondence scores, with its four modes explicitly encoding the probe row, probe column, gallery row, and gallery column. This construction differs fundamentally from reshaping a pooled $512$-dimensional embedding: the tensor modes retain the spatial semantics of the two images and therefore expose a natural factorization over local correspondences.

The initial Born-rule TT classifier was consistently weaker than the corresponding linear models; we therefore use the linear TT/MPS family as the primary structured model and retain Born-rule variants as controls. We evaluate three tensor orderings:
\begin{align}
\pi_{\mathrm{group}}
  &= (y_A,x_A,y_B,x_B) \\
\pi_{\mathrm{inter}}
  &= (y_A,y_B,x_A,x_B) \\
\pi_{\mathrm{rel}}
  &= (y_A,x_A,\Delta y,\Delta x)
\end{align}
where $\Delta y$ and $\Delta x$ encode relative probe--gallery spatial offsets. The grouped ordering preserves the within-image spatial coordinates, the interleaved ordering places corresponding probe and gallery coordinates adjacent, and the relative ordering explicitly represents translation-invariant spatial relationships.
Primary TT/MPS models are constrained to satisfy pair symmetry,
$s(A,B)=s(B,A)$
so that the learned similarity is invariant to exchanging probe and gallery images. CP and Tucker factorizations are included as additional controls to determine whether any observed advantage arises from generic low-rank tensor approximation rather than from the sequential factorization imposed by TT/MPS.



\subsection{Development Protocol}

CASIA-WebFace~\cite{yi2014casia} is used exclusively for model fitting and model selection. For each of five random seeds, we use 400 training identities and 100 identity-disjoint validation identities, with at most 20 images per identity. This yields 40,000 training pairs and 20,000 validation pairs per seed. All design choices, including tensor ordering, bond dimension, and model family selection, are determined solely by this development protocol.

LFW~\cite{huang2007lfw}, CFP-FP~\cite{sengupta2016cfp}, and CPLFW~\cite{zheng2018cplfw} are reserved as external biometric evaluation protocols and are never used for model or tensor-structure selection. We report AUC, EER, fold-calibrated accuracy, and TAR at $\mathrm{FAR}\in\{10^{-2},10^{-3}\}$. As a protocol integrity check, each evaluation fold must contain both genuine and impostor pairs. This check detected a mirror-specific CPLFW pair-ordering error during development and prevents class-pure folds from silently yielding undefined (NaN) metrics.

\subsection{Parameter Efficiency and External AUC}

\begin{figure}[t]
\centering
\includegraphics[width=.90\textwidth]{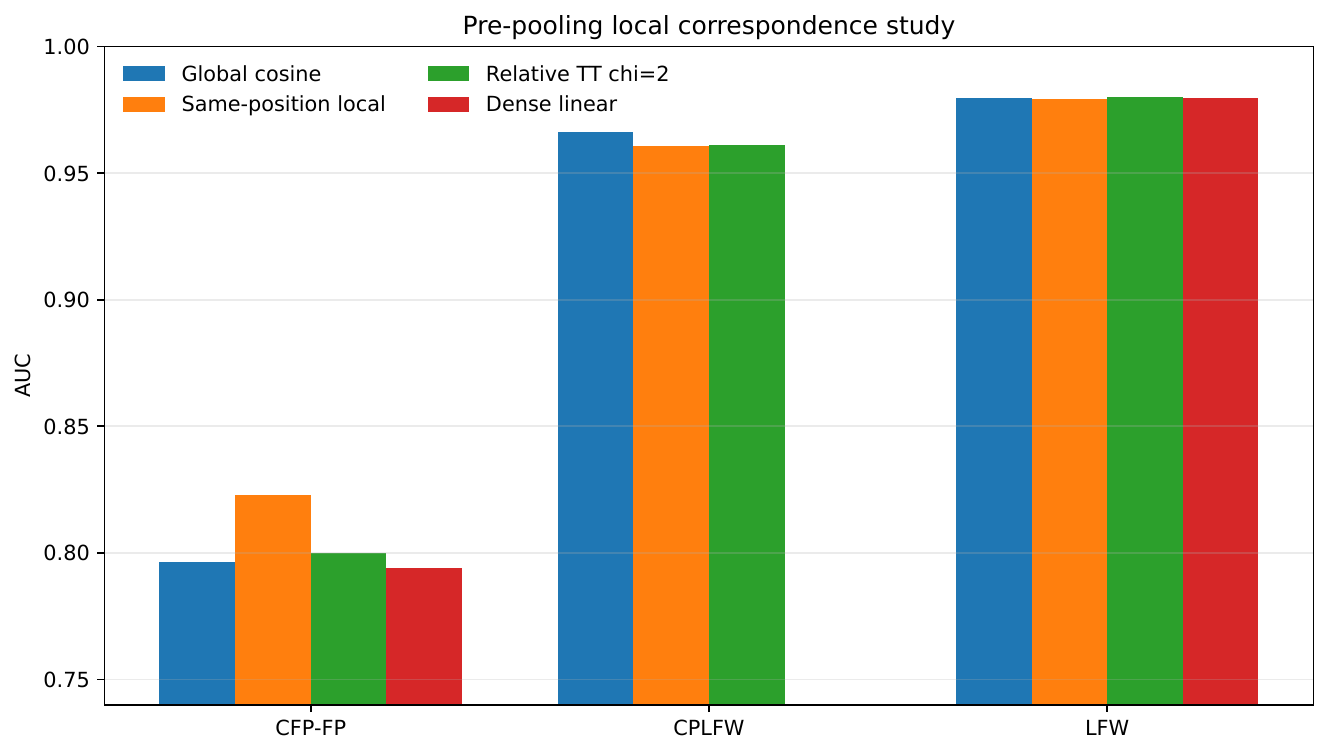}
\caption{External AUC for the pre-pooling correspondence study. Local evidence is especially useful on CFP-FP, but the parameter-free same-position matcher remains strongest there.}
\label{fig:localauc}
\end{figure}

\begin{table}[t]
\centering
\caption{External AUC and TAR at FAR=$10^{-3}$ for key correspondence methods.}
\label{tab:local}
\begin{adjustbox}{max width=\textwidth}
\begin{tabular}{llcc}
\toprule
\textbf{Dataset} & \textbf{Method} & \textbf{AUC} $\uparrow$ & \textbf{TAR@$10^{-3}$ $\uparrow$}\\
\midrule
\multirow{4}{*}{CFP-FP}
& Global cosine & 0.796314 & \textbf{0.173429}\\
& Same-position local & \textbf{0.822771} & 0.155714\\
& Relative TT $\chi=2$ & 0.800044 & 0.010229\\
& Dense linear & 0.794068 & 0.038629\\
\midrule
\multirow{3}{*}{CPLFW}
& Global cosine & \textbf{0.966074} & \textbf{0.873667}\\
& Same-position local & 0.960581 & 0.743667\\
& Relative TT $\chi=2$ & 0.961140 & 0.780533\\
\midrule
\multirow{3}{*}{LFW}
& Global cosine & 0.979789 & 0.950667\\
& Same-position local & 0.979403 & \textbf{0.951333}\\
& Relative TT $\chi=2$ & \textbf{0.980201} & 0.951000\\
\bottomrule
\end{tabular}
\end{adjustbox}
\end{table}

The model selected exclusively on the CASIA validation set is the relative-coordinate, pair-symmetric TT with bond dimension $\chi=2$. It contains only 122 trainable parameters, corresponding to 5.1\% of the 2,403 parameters in the dense linear correspondence head. As shown in Fig.~\ref{fig:localauc} and Table~\ref{tab:local}, tensorization at the semantically structured local-correspondence interface is substantially more favorable than tensorization of the global compatibility operator. On LFW, the selected TT achieves an AUC of 0.980201, marginally exceeding the global cosine baseline. On CFP-FP, it reaches 0.800044, outperforming both global cosine and the dense linear correspondence head, making it the strongest \emph{learned} correspondence-based matcher among the evaluated methods. Nevertheless, the parameter-free same-position local score achieves a higher AUC of 0.822771, demonstrating that part of the performance gain arises from the underlying local spatial correspondence itself rather than from the learned TT parameterization. On CPLFW, the global cosine baseline remains the strongest method, indicating that the advantage of semantic tensorization is protocol-dependent and does not uniformly translate across pose-variation conditions.

\subsection{Low-FAR Performance Reveals Operational Limitations}
\begin{figure}[t]
\centering
\includegraphics[width=.96\textwidth]{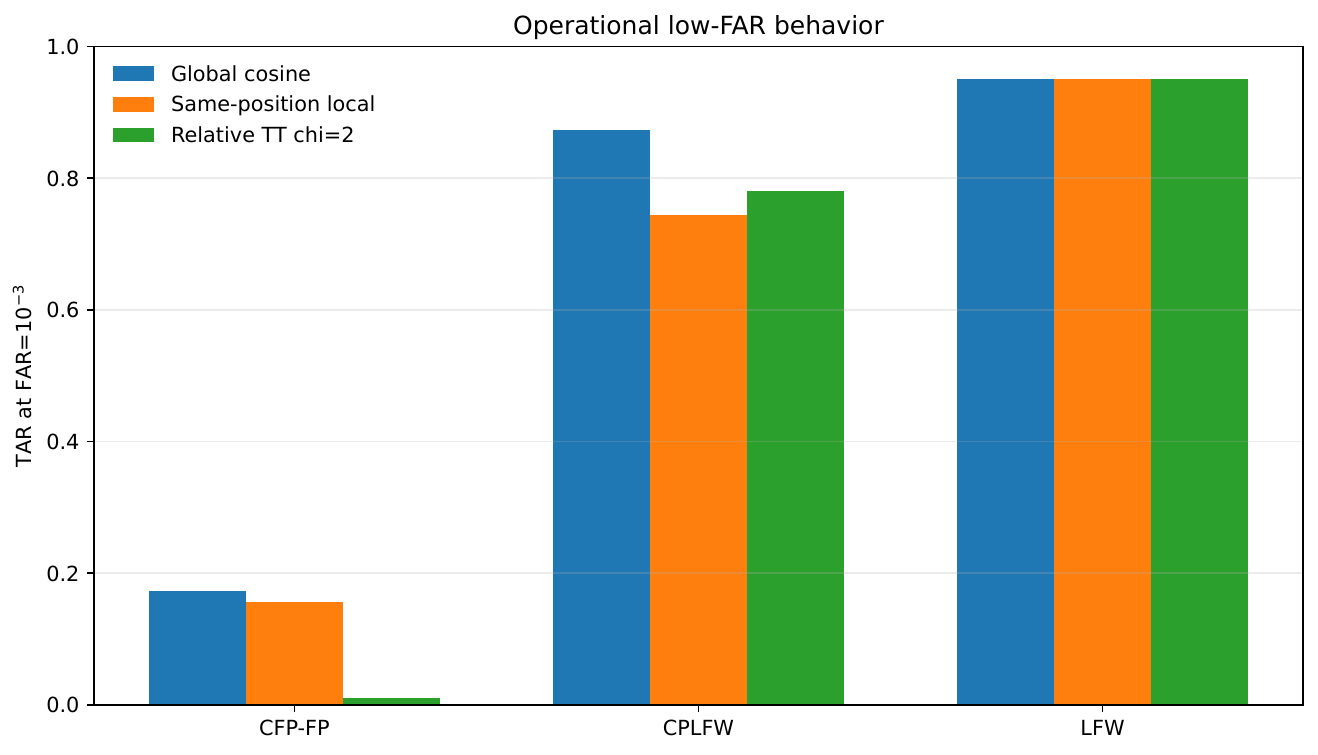}
\caption{Operational TAR at FAR=$10^{-3}$. The TT preserves LFW but loses a substantial acceptance rate on the two pose-stress protocols.}
\label{fig:lowfar}
\end{figure}

The most consequential limitation emerges at the operational biometric threshold rather than in aggregate AUC. On CFP-FP, the TT achieves a TAR of only 0.010229 at $\mathrm{FAR}=10^{-3}$, compared with 0.173429 for global cosine; on CPLFW, the corresponding TARs are 0.780533 and 0.873667, respectively. Thus, as shown in Fig.~\ref{fig:lowfar}, an AUC-only assessment would give an overly favorable impression of the TT model. Although the relative-coordinate TT provides an effective and highly parameter-efficient compression of the local correspondence tensor and offers a plausible geometry-aware representation, its unfavourable impostor-tail behaviour prevents it from being considered an operationally superior pose-robust verifier. These results emphasize that biometric model evaluation must include low-FAR operating points, where tail behaviour can differ substantially from aggregate ranking metrics such as AUC.


\subsection{Structural Diagnostics and Failure Attribution}
Our experiments are designed so that each negative result has a diagnostic interpretation rather than merely a lower score. Table~\ref{tab:diagnostics} summarizes the structural assumptions and the evidence used to accept or reject them. In compatibility alignment, the decisive observation is not simply that MPO is worse than Ridge. The rank-64 SVD control proves that the operator contains compressible structure, while the much larger loss of MPO at a comparable storage scale indicates that the missing property is \emph{tensor locality}. The operator-Schmidt entropy analysis then supplies a mechanism: important bipartitions remain close to maximally entangled under the tested site orderings and basis changes.

In local correspondence, the selected relative-coordinate TT gives the opposite structural signal. It is chosen independently on the CASIA validation set, uses only 122 parameters, and slightly exceeds dense linear matching on the two external AUC settings where both are available. This shows that a low-bond factorization can exploit the spatial object. Yet the same model fails badly at CFP-FP low FAR, so structural compressibility and biometric calibration are separate properties. A method can have the right inductive bias for average ranking while still reshaping the score tails in a harmful way.

\begin{table}[t]
\centering
\caption{Structural assumptions and the controls used to test them.}
\label{tab:diagnostics}
\begin{adjustbox}{max width=\textwidth}
\begin{tabular}{p{2.4cm}p{3.8cm}p{4.8cm}}
\toprule
\textbf{Interface} & \textbf{Hypothesis} & \textbf{Diagnostic evidence}\\
\midrule
Compatibility map & A 512$\times$512 map is low-bond after tensorization & SVD compresses moderately, but TT/MPO remains much worse; operator-Schmidt entropies stay high across tested bases.\\
Local correspondence & Spatial probe/gallery axes support low-bond matching & Relative $\chi=2$ TT is selected on disjoint validation and gives about $20\times$ matcher compression with competitive AUC.\\
Operational verification & Better AUC implies a better biometric operating point & Rejected: CFP-FP TT AUC is competitive while TAR@$10^{-3}$ collapses.\\
\bottomrule
\end{tabular}
\end{adjustbox}
\end{table}

\section{Cross-Placement Analysis}

\subsection{Arbitrary Versus Semantic Tensorization}
The two experiments differ primarily in the structure exposed to the same TN family. Compatibility alignment is a global operator whose tensor ordering is imposed after representation learning; its high operator Schmidt entropy indicates substantial cross-site dependence that low-bond TT/MPO factorizations cannot efficiently preserve. In contrast, local correspondence is naturally a four-index geometric tensor, and relative coordinates align the factorization with spatial displacement structure.
This structural difference explains why semantic tensorization enables strong compression whereas arbitrary tensorization does not. However, structural fit is not sufficient for biometric superiority: the parameter-free local matcher still achieves higher CFP-FP AUC, and the learned TT degrades low-FAR performance. Thus, semantic tensorization is a prerequisite for effective TN compression, not a guarantee of operational robustness.

\subsection{Matched Controls as a Placement Test}
A useful quantum-inspired claim must survive simpler explanations. In compatibility alignment, truncated SVD investigate whether the ordinary matrix rank already explains the available compression.  In local correspondence, CP/Tucker and dense linear heads investigate whether generic low rank or parameter count explains the result, while same-position, nearest-patch, and optimal-transport scores investigate whether learning is necessary at all. The resulting hierarchy prevents compactness from being mistaken for superiority.

\subsection{Why Biometric Tails Matter for Quantum Vision}
Face verification differs from balanced image classification because it is often deployed at very low false-accept rates. A method can improve average ranking while moving a small number of impostors into the high-score tail. The P3 result demonstrates this directly. This issue is likely to be even more important for finite-shot or noisy quantum models, where measurement variation can perturb score tails before AUC changes appreciably. Quantum biometric work should therefore report TAR@FAR and calibration alongside AUC or accuracy.

\subsection{Implications Beyond Face Recognition}
The placement principle extends to other quantum-CV pipelines built on pretrained vision features. CNN feature maps, ViT token sequences, optical-flow fields, and patch-correlation volumes have different natural axes and factorization structures; pooled embeddings often do not. Before introducing an MPS, MPO, or qubit register, one should therefore specify the semantic meaning of each site and justify why the chosen ordering should yield compressible correlations.

This also calls for stronger evaluation of ``quantum-inspired compression.'' Parameter count alone is insufficient, as SVD, CP/Tucker, separable layers, or low-rank adapters may provide comparable compression. A meaningful quantum-inspired claim requires that the proposed product structure matches the task and preserves performance at a favorable resource budget against these simpler controls. Our two experiments demonstrate both outcomes within the same biometric setting.

\section{Placement Doctrine for Quantum Computer Vision}
Our experiments suggest four practical principles: i)Start from the interface, not the circuit: Identify the available information and its natural factorization before selecting a quantum or TN primitive. ii) Test structural assumptions directly: For an MPO, compare against SVD and examine operator entanglement; for a local tensor, compare against parameter free local scores and alternative low-rank factorizations. iii) Separate compression from task performance: A $20\times$ smaller matcher is a meaningful compression result even without an accuracy gain; the claim should remain appropriately limited to compression. iv) Evaluate operational performance: External identity-disjoint protocols and low FAR TAR, since these can reveal failures hidden by aggregate AUC.

These principles extend beyond face recognition to quantum-CV pipelines based on CNN features, token sequences, and attention maps. The key question is not whether a representation can be reshaped into a tensor, but whether its factorization reflects the task's semantics and correlation structure.



\section{Limitations and Reproducibility}
Both experiments are quantum-inspired classical studies and do not involve execution on quantum hardware. The compatibility experiment evaluates TN/MPO representations of a cross-model face-space adapter, while the local experiment evaluates TT/MPS matchers over spatial correspondence tensors. Accordingly, we make no claims regarding quantum speedup, quantum advantage, or hardware efficiency. The two studies also use different face representations and are not intended as a single jointly optimized end-to-end system. Their purpose is instead to provide a controlled comparison of alternative tensorization placements. The local study provides the stronger statistical evaluation, using five CASIA identity-disjoint splits, external LFW/CFP-FP/CPLFW protocols, fold-calibrated metrics, paired sign-flip tests, and hierarchical bootstrap comparisons. Future work should extend the placement doctrine to template-based IJB protocols, face-quality and uncertainty modeling, and trainable quantum encodings, while maintaining independent evaluation to avoid turning previously observed results into new placement claims.



\section{Conclusion}

QFacePlacement investigates when quantum-inspired factorization is structurally justified for modern face verification. Tensorizing a global $512$-D compatibility operator is straightforward but structurally mismatched: deterministic MPO/TT compression remains substantially weaker than dense Ridge and storage-matched SVD. In contrast, moving the tensorization point before global pooling exposes a genuine probe--gallery correspondence tensor. A relative-coordinate, pair-symmetric TT then achieves approximately $20\times$ parameter compression while preserving LFW performance and maintaining competitive aggregate AUC. However, low-FAR evaluation shows that this compression does not translate into operational superiority under cross-pose verification.

The central lesson is therefore \emph{placement before architecture}. Quantum-inspired models are most credible when their factorization follows task semantics, demonstrates compression beyond simple low-rank controls, and preserves performance at operationally relevant thresholds. The positive and negative results provide a practical boundary for structured quantum computer vision: rather than tensorizing already-optimized representations arbitrarily, identify interfaces where the task itself exposes a meaningful and compressible factorization.

\bibliographystyle{splncs04}
\bibliography{references}

@inproceedings{deng2019arcface,
  title={ArcFace: Additive Angular Margin Loss for Deep Face Recognition},
  author={Deng, Jiankang and Guo, Jia and Xue, Niannan and Zafeiriou, Stefanos},
  booktitle={CVPR}, year={2019}
}

@inproceedings{kim2022adaface,
  title={AdaFace: Quality Adaptive Margin for Face Recognition},
  author={Kim, Minchul and Jain, Anil K. and Liu, Xiaoming},
  booktitle={CVPR}, year={2022}
}

@inproceedings{schroff2015facenet,
  title={FaceNet: A Unified Embedding for Face Recognition and Clustering},
  author={Schroff, Florian and Kalenichenko, Dmitry and Philbin, James},
  booktitle={CVPR}, year={2015}
}

@inproceedings{wang2018cosface,
  title={CosFace: Large Margin Cosine Loss for Deep Face Recognition},
  author={Wang, Hao and Wang, Yitong and Zhou, Zheng and Ji, Xing and Gong, Dihong and Zhou, Jingchao and Li, Zhifeng and Liu, Wei},
  booktitle={CVPR}, year={2018}
}

@inproceedings{meng2021magface,
  title={MagFace: A Universal Representation for Face Recognition and Quality Assessment},
  author={Meng, Qiang and Zhao, Shichao and Huang, Zhida and Zhou, Feng},
  booktitle={CVPR}, year={2021}
}

@techreport{huang2007lfw,
  title={Labeled Faces in the Wild: A Database for Studying Face Recognition in Unconstrained Environments},
  author={Huang, Gary B. and Ramesh, Manu and Berg, Tamara and Learned-Miller, Erik},
  institution={University of Massachusetts Amherst}, year={2007}
}

@inproceedings{sengupta2016cfp,
  title={Frontal to Profile Face Verification in the Wild},
  author={Sengupta, Soumyadip and Chen, Jun-Cheng and Castillo, Carlos and Patel, Vishal M. and Chellappa, Rama and Jacobs, David W.},
  booktitle={WACV}, year={2016}
}

@article{zheng2018cplfw,
  title={Cross-Pose LFW: A Database for Studying Cross-Pose Face Recognition in Unconstrained Environments},
  author={Zheng, Tianyue and Deng, Weihong},
  journal={Beijing University of Posts and Telecommunications Technical Report}, year={2018}
}

@article{yi2014casia,
  title={Learning Face Representation from Scratch},
  author={Yi, Dong and Lei, Zhen and Liao, Shengcai and Li, Stan Z.},
  journal={arXiv preprint arXiv:1411.7923}, year={2014}
}

@inproceedings{phan2022deepfaceemd,
  title={DeepFace-EMD: Re-Ranking Using Patch-Wise Earth Mover's Distance Improves Out-of-Distribution Face Identification},
  author={Phan, Huy and Yin, Yutong and Manley, Thomas and others},
  booktitle={CVPR}, year={2022}
}

@article{schuld2019featurehilbert,
  title={Quantum Machine Learning in Feature Hilbert Spaces},
  author={Schuld, Maria and Killoran, Nathan},
  journal={Physical Review Letters}, volume={122}, number={4}, pages={040504}, year={2019}
}

@article{havlicek2019supervised,
  title={Supervised Learning with Quantum-Enhanced Feature Spaces},
  author={Havl{\'i}{\v c}ek, Vojt{\v e}ch and C{\'o}rcoles, Antonio D. and Temme, Kristan and Harrow, Aram W. and Kandala, Abhinav and Chow, Jerry M. and Gambetta, Jay M.},
  journal={Nature}, volume={567}, pages={209--212}, year={2019}
}

@article{mcclean2018barren,
  title={Barren Plateaus in Quantum Neural Network Training Landscapes},
  author={McClean, Jarrod R. and Boixo, Sergio and Smelyanskiy, Vadim N. and Babbush, Ryan and Neven, Hartmut},
  journal={Nature Communications}, volume={9}, pages={4812}, year={2018}
}

@article{perezsalinas2020datareuploading,
  title={Data Re-uploading for a Universal Quantum Classifier},
  author={P{\'e}rez-Salinas, Adri{\'a}n and Cervera-Lierta, Alba and Gil-Fuster, Elies and Latorre, Jos{\'e} I.},
  journal={Quantum}, volume={4}, pages={226}, year={2020}
}

@article{zhu2024mgqcnn,
  title={Quantum Face Recognition with Multigate Quantum Convolutional Neural Network},
  author={Zhu, X. and Bouridane, A. and Celebi, M. E. and Konar, D. and Angelov, P. and Ni, Q. and Jiang, R.},
  journal={IEEE Transactions on Artificial Intelligence}, volume={5}, number={12}, pages={6330--6341}, year={2024}
}

@inproceedings{bhatia2026nearterm,
  title={Evaluating Near-Term Quantum Machine Learning for Face Recognition},
  author={Bhatia, Krishna and Chandaliya, Praveen Kumar and Devendrababu, S.},
  booktitle={AI for Forensics and Biometric Applications}, publisher={Springer}, year={2026}
}

@incollection{gauba2026attention,
  title={Quantum-Inspired Attention for Efficient Face Recognition},
  author={Gauba, A.}, booktitle={Quantum AI and NLP}, series={CCIS}, volume={2659}, pages={58--66}, publisher={Springer}, year={2026}
}

@article{salari2023ghost,
  title={Quantum Face Recognition Protocol with Ghost Imaging},
  author={Salari, V. and others}, journal={Scientific Reports}, volume={13}, pages={2401}, year={2023}
}

@article{singh2025eigenface,
  title={Quantum-Enhanced Eigenface Algorithm for Face Verification},
  author={Singh, S. and Thakur, A. and Sinha, D. K. and Sinha, G. and Hasan, M.},
  journal={Journal of Information Systems Engineering and Management}, volume={10}, number={36s}, pages={588--603}, year={2025}
}

@inproceedings{singh2025qpca,
  title={Accelerating Face Biometric Verification via Quantum PCA and Hybrid Processing},
  author={Singh, S. and Thakur, A. and Hussain, M. I. and Hasan, M.}, booktitle={ICIRCA}, year={2025}
}

@article{oseledets2011tt,
  title={Tensor-Train Decomposition}, author={Oseledets, Ivan V.}, journal={SIAM Journal on Scientific Computing}, volume={33}, number={5}, pages={2295--2317}, year={2011}
}

@article{stoudenmire2016supervised,
  title={Supervised Learning with Tensor Networks}, author={Stoudenmire, E. Miles and Schwab, David J.}, journal={Advances in Neural Information Processing Systems}, volume={29}, year={2016}
}

@inproceedings{novikov2015tensorizing,
  title={Tensorizing Neural Networks}, author={Novikov, Alexander and Podoprikhin, Dmitrii and Osokin, Anton and Vetrov, Dmitry}, booktitle={NeurIPS}, year={2015}
}

@article{schonemann1966procrustes,
  title={A Generalized Solution of the Orthogonal Procrustes Problem}, author={Sch{\"o}nemann, Peter H.}, journal={Psychometrika}, volume={31}, pages={1--10}, year={1966}
}
\end{document}